\documentclass[final,1p,times]{elsarticle}

\usepackage{booktabs}
\usepackage{hyperref}

\makeatletter
\def\ps@pprintTitle{%
 \let\@oddhead\@empty
 \let\@evenhead\@empty
 \def\@oddfoot{}%
 \let\@evenfoot\@oddfoot}
\makeatother

\usepackage{amsmath}
\usepackage{graphicx}
\usepackage{amssymb}
\usepackage{commath}
\usepackage{mathtools}
\usepackage{subcaption}
\usepackage{float}
\usepackage{multirow}

\newcommand{\diagdots}[3][-25]{%
  \rotatebox{#1}{\makebox[0pt]{\makebox[#2]{\xleaders\hbox{$\cdot$\hskip#3}\hfill\kern0pt}}}%
}

\DeclareMathOperator*{\argmin}{arg\,min}
\DeclareMathOperator*{\E}{\mathbb{E}}

\begin{document}

\begin{frontmatter}

\title{Exact Backpropagation in Binary Weighted Networks with Group Weight Transformations}
\author{Yaniv Shulman}
\address{yaniv@aleph-zero.info}

\begin{abstract}
Quantization based model compression serves as high performing and fast approach for inference that yields models which are highly compressed when compared to their full-precision floating point counterparts. The most extreme quantization is a 1-bit representation of parameters such that they have only two possible values, typically -1(0) or +1. Models that constrain the weights to binary values enable efficient implementation of the ubiquitous dot product using only additions without requiring floating point multiplications which is beneficial for resource-constrained inference. The main contribution of this work is the introduction of a method to smooth the combinatorial problem of determining a binary vector of weights to minimize the expected loss for a given objective by means of empirical risk minimization with backpropagation. This is achieved by approximating a multivariate binary state over the weights utilizing a deterministic and differentiable transformation of real-valued, continuous parameters. The proposed method adds little overhead in training, can be readily applied without any substantial modifications to the original architecture, does not introduce additional saturating nonlinearities or auxiliary losses, and does not prohibit applying other methods for binarizing the activations. Contrary to common assertions made in the literature, it is demonstrated that binary weighted networks can train well with the same standard optimization techniques and similar hyperparameter settings as their full-precision counterparts, specifically momentum SGD with large learning rates and $L_2$ regularization. To conclude experiments demonstrate the method performs remarkably well across a number of inductive image classification tasks with various architectures compared to their full-precision counterparts. The source code is publicly available at \url{https://bitbucket.org/YanivShu/binary_weighted_networks_public}.

\end{abstract}
\end{frontmatter}

\section{Introduction}
\label{s:Introduction}
Contemporary artificial neural networks (ANN) have achieved state-of-the-art results in a multitude of learning tasks. Often these models include millions of parameters which form dense structures enabling efficient parallel computing by utilizing specialized software and hardware. However the dependency of these models on substantial hardware resources limits their utility on resource constrained hardware such as mobile and low power embedded devices. One approach to reduce computational resources is \emph{model compression} that transforms an initial cumbersome architecture into a more efficient architecture that require less space and compute resources while minimizing performance loss to an acceptable degree. Model compression is typically achieved by reducing the number of parameters in the model and/or by quantizing the parameters and activations so that they use less bits to encode the data flowing through the network. 

There are many approaches suggested for reducing the number of parameters including weight pruning \cite{10.5555/2969239.2969366}, architecture learning \cite{DBLP:conf/bmvc/SrinivasB16}, distilling knowledge \cite{44873}, structured pruning \cite{NIPS2016_41bfd20a} and $L_0$ regularization \cite{louizos2018learning, shulman2020diffprune}. The interested reader is referred to \cite{DBLP:conf/mlsys/BlalockOFG20, journals/corr/abs-1710-09282} for recent reviews.

Quantization based model compression serves as high performing and fast approach for inference that yields highly compressed models compared to their full-precision floating point counterparts. The most extreme quantization is a 1-bit representation of parameters and activations such that they have only two possible values, typically -1(0) or +1. An ANN that is restricted to binary representations is typically known as a Binary Neural Network (BNN). Models that constrain the weights to binary values enable efficient implementation of the ubiquitous dot product using only additions without requiring floating point multiplications. Furthermore networks which restrict both weights and activations to binary values enable significant computational acceleration in inference by utilizing highly efficient bitwise XNOR and Bitcount operations that can be further optimized in specialized hardware. Therefore these models are an attractive alternative to full-precision ANNs where power efficiency and constrained compute resources are important considerations \cite{Qin_2020, NIPS2015_3e15cc11, DBLP:conf/eccv/RastegariORF16, electronics8060661}. Another compelling approach is to binarize only parts of the network that benefit the most from the quantization and keep other layers at high precision. In fact most proposed BNNs use partial binarization since typically at least the fully connected output layer and the first convolution layer weights are kept at a higher precision \cite{electronics8060661, Qin_2020}. Additional examples include retaining the parameters of the batch normalization layers at high precision \cite{10.1007/978-3-030-01237-3_23}, apply a scaling factor to the binary weights \cite{DBLP:conf/eccv/RastegariORF16, DBLP:journals/corr/abs-1909-13863, Martinez2020Training, SakrCWGS18} or floating point parametrized activations \cite{DBLP:journals/corr/abs-1904-05868, Martinez2020Training}.

Many learning algorithms and in particular neural networks typically employ gradient-based optimizers such as the Backpropagation algorithm \cite{Rumelhart:1986we}. Models that are designed to have a continuous relationship between parameters and the training objective enable the computation of exact gradients which in turn enable efficient
optimization \cite{journals/corr/BengioLC13}. Many of the existing methods in the literature for ANN quantization such as \cite{NIPS2016_d8330f85, Liu_2018_ECCV, Cai_2017_CVPR, DBLP:journals/corr/abs-1812-11800, DBLP:conf/cvpr/QinGLSWYS20} employ non-differentiable quantization techniques that require the use of gradient estimators resulting in divergence between the forward pass and backpropagation and therefore decreased training efficacy \cite{NIPS2017_1c303b0e}. The challenge is then combining discrete valued weights for which the gradient is undefined with the effective backpropagation method for training neural networks.

The main contribution of this work is the introduction of a method to smooth the combinatorial problem of finding a binary vector of weights to minimize the expected loss for a given objective by means of empirical risk minimization with backpropagation. This is achieved by approximating a multivariate binary state over the weights utilizing a deterministic and differentiable transformation of real-valued, continuous parameters. The proposed method adds little overhead in training, can be readily applied without any modifications to the original architecture, does not introduce additional saturating nonlinearities or auxilary losses and does not prohibit applying other methods for binarizing the activations. Contrary to common assertions made in the literature, it is demonstrated that binary weighted networks can train well with the same standard optimization techniques and similar hyperparameter settings as their full-precision counterparts, specifically momentum SGD with large learning rates and $L_2$ regularization \cite{Qin_2020}. To conclude experiments demonstrate little and even a modest gain in accuracy for a number of inductive image classification tasks compared to their full-precision counterparts. The source code is publicly available at \url{https://bitbucket.org/YanivShu/binary_weighted_networks_public}.

Note the term \emph{differentiable} is used in this paper in the context of training neural networks, i.e. allowing a small number of points where the first order derivatives do not exist. A common example is the use of rectifiers in the calculation graph such as the Relu activation \cite{NairH10}.

\section{Proposed method}
\label{S:Proposed method}
\subsection{Binary group weight transformations}
\label{S:Binary group weight transformations}
Let $\boldsymbol{\pi} \in \{-1, 1 \} ^{\abs{\boldsymbol{\pi}}}$ be the binary valued weights (parameters) of a hypothesis $h(\cdot,\boldsymbol{\pi}): X \rightarrow Y$ such as a binary weighted neural network where $\abs{\boldsymbol{\pi}}$ denotes the cardinality of $\boldsymbol{\pi}$. Let $\mathcal{D}$ be a training set consisting of $N$ i.i.d. instances $\{(x_1,y_1),\ldots, (x_N,y_N)\}$. The empirical risk $\mathcal{R}$ associated with the hypothesis $h(\cdot,\boldsymbol{\pi})$ is defined as:

\begin{align}
\mathcal{R}_h(\boldsymbol{\pi}) &= \frac{1}{N} \left(\sum_{i=1}^{N}\mathcal{L} \left( h(x_i;\boldsymbol{\pi}),y_i \right) \right) \label{eq:risk_pi} \\
\boldsymbol{\pi}^{*} &= \argmin_{\boldsymbol{\pi}} \mathcal{R}_h(\boldsymbol{\pi}) \label{eq:risk_objective}
\end{align}

Where $\boldsymbol{\pi}$ is constrained to take values in $\{-1,1\}$ and $\mathcal{L}:Y \times Y \rightarrow \mathbb{R}_{\geq 0}$ is a loss function that measures the discrepancy between the true value $y_i$ and the predicted outcome $\hat{y}_i = h(x_i;\boldsymbol{\pi})$. The goal of the optimization problem is to find $\boldsymbol{\pi}^{*}$ given the hypothesis $h$ and data $\mathcal{D}$ for which the empirical risk $\mathcal{R}_h(\boldsymbol{\pi})$ is minimal.

Minimizing the objective \eqref{eq:risk_pi} provably is a hard combinatorial problem with complexity exponential in respect to $\abs{\boldsymbol{\pi}}$. Alternative methods such as gradient based optimization cannot be readily used due to $\mathcal{R}_{h}(\boldsymbol{\pi})$ not being differentiable w.r.t. $\boldsymbol{\pi}$. To overcome this challenge a deterministic differentiable relaxation of the hard binary constraints governing $\boldsymbol{\pi}$ is proposed that enables solving a surrogate minimization problem efficiently and deterministically using common gradient based optimizers. To enable efficient backpropagation during training the hard constraint of the weights $\boldsymbol{\pi}$ being exactly binary may be relaxed and replaced with a soft constraint of being approximately one or negative one. Let $\boldsymbol{\phi} \in \mathbb{R}^{\abs{\boldsymbol{\pi}}} $ be a real valued vector and $u(\cdot): \mathbb{R} \rightarrow [-1,1]$ be a differentiable function from the real numbers to the range $[-1, 1]$ e.g. the hyperbolic tangent $tanh(\cdot)$. Equations \eqref{eq:eff_a} - \eqref{eq:eff_e} define a deterministic and differentiable transformation $g(\cdot, \cdot)$ that maps vectors in $\mathbb{R}^{\abs{\boldsymbol{\pi}}}$ to be approximately binary i.e. $ g \left( \boldsymbol{\phi}, \zeta \right) \in \{ w \, | \, \abs{w} - 1 < \epsilon \} $ for some small $\epsilon \in \mathbb{R}$.
\begin{align}
\mathbf{w}^+ &= \{ u \left( \phi_k \right) \, | \, u \left( \phi_k \right) > 0 \, , \: k= 1,\ldots , \abs{\boldsymbol{\phi}} \} \label{eq:eff_a} \\
\mathbf{w}^- &= \{ u \left( \phi_k \right) \, | \, u \left( \phi_k \right) \leq 0 \, , \: k= 1,\ldots , \abs{\boldsymbol{\phi}} \} \label{eq:eff_b} \\
\mathbf{w}^1 &= \left( \mathbf{w}^+ - \bar{w}^+ \right) e^{-\zeta} + 1 \label{eq:eff_c} \\
\mathbf{w}^{-1} &= \left( \mathbf{w}^- - \bar{w}^- \right) e^{-\zeta} - 1 \label{eq:eff_d} \\
\mathbf{w} &= \mathbf{w}^1 \cup \mathbf{w}^{-1} \label{eq:eff_e}
\end{align}

Where $\phi_k$ denotes the $k$-th element of $\boldsymbol{\phi}$; $\bar{w}^+$ and $\bar{w}^-$ are the mean of $\mathbf{w}^+$ and $\mathbf{w}^-$ respectively; and $\zeta \in \mathbb{R}_{\geq 0}$. The transformation defined by $g(\cdot, \cdot)$ conceptually comprises $\mathbf{w}$ of two partitions: $\mathbf{w}^{-1}$ and $\mathbf{w}^1$, such that by definition under the assumption that $\abs{\mathbf{w}^{-1}} > 0$ and $\abs{\mathbf{w}^1} > 0$ then $\E ( \mathbf{w}^{-1} ) = -1$ and $\E ( \mathbf{w}^{1} ) = 1$. The variance of both $\mathbf{w}^1$ and $\mathbf{w}^{-1}$ is controlled by $\zeta$ and since $ \abs{w_k^+ - w_l^+} < 1 $ and $ \abs{w_k^- - w_l^-} < 1 $ it may be set as small as practically useful and therefore $\mathbf{w}$ is exactly binary in the limit when $\zeta \rightarrow \infty$. Note the gradient of $\mathbf{w}$ w.r.t. $\boldsymbol{\phi}$ is non-degenerate provided that $ 2 \leq \abs{\mathbf{w}^1} \leq \abs{\mathbf{w}} - 2 $ i.e. there are at least two members in each of $\mathbf{w}^+$ and $\mathbf{w}^-$.

Having defined $g(\cdot, \cdot)$, reconsider the hypothesis $h$ and associated empirical risk $\mathcal{R}_h$ following reparameterization of $\boldsymbol{\pi}$ given a partition of $\boldsymbol{\pi}$ to $M$ subsets $\boldsymbol{\pi}_1 \, , \ldots  \, , \boldsymbol{\pi}_M$:

\begin{align}
\boldsymbol{\pi} = \lim_{\zeta \to \infty} g & \left( \boldsymbol{\phi}, \zeta \right) \, , \quad \boldsymbol{\pi} = \bigcup\limits_{j=1}^{M} \boldsymbol{\pi}_j \, , \quad \boldsymbol{\phi} = \bigcup\limits_{j=1}^{M} \boldsymbol{\phi}_j \, , \quad \abs{\boldsymbol{\pi}_j} \geq 2 \, , \label{eq:risk_phi_a} \\
\lim_{\zeta \to \infty} \mathcal{R}_{h}(\boldsymbol{\phi}, \zeta) &= \lim_{\zeta \to \infty} \frac{1}{N} \left(\sum_{i=1}^{N}\mathcal{L} \left( h(x_i;\cup_{j=1}^{M} \, g \left( \boldsymbol{\phi}_j, \zeta \right),y_i \right) \right) \label{eq:risk_phi_b} \\
\boldsymbol{\phi}^{*} &= \argmin_{\boldsymbol{\phi}} \lim_{\zeta \to \infty} \mathcal{R}_{h} (\boldsymbol{\phi}, \zeta) \label{eq:risk_phi_c}
\end{align}

The objectives in equations \eqref{eq:risk_pi} and \eqref{eq:risk_phi_b} are equivalent in the limit as $\zeta \rightarrow \infty$. However for reasonably low values of $\zeta$ the formulation in equations \eqref{eq:risk_phi_a} - \eqref{eq:risk_phi_c} can be used as a differentiable surrogate to the objective in equation \eqref{eq:risk_pi} due to replacement of the binary weights $\boldsymbol{\pi}$ with the smoothed approximate binary weights $\mathbf{w}$.  Subsequently this enables the use of gradient based optimizers to find an approximate solution to the original hard combinatorial problem with low quantization error.

\subsection{Reduction of quantization error with $L_2$ regularization}
The inclusion of $u(\cdot): \mathbb{R} \rightarrow [-1,1]$ in equations \eqref{eq:eff_a} and \eqref{eq:eff_b} enables theoretical bounds on the divergence of the binarized weights $\mathbf{w}$ from $\pm 1$ respectively. The inclusion of such nonlinearities is a common approach and often the hyperbolic tangent is used for this purpose in training BNNs \cite{DBLP:journals/corr/abs-1904-05868, Martinez2020Training,  Gong_2019_ICCV, lahoud2019selfbinarizing, DBLP:conf/cvpr/QinGLSWYS20} or the hard tanh and its variants \cite{NIPS2016_d8330f85, SakrCWGS18}. The inclusion of superfluous saturating nonlinearities changes the objective in a non-trivial way and slows training as these typically have substantial areas of their domain where gradients are very small or practically zero. To mitigate these shortcomings it is proposed that in practice the function $u(\cdot): \mathbb{R} \rightarrow [-1,1]$ is removed and instead soft constraints are introduced on $\boldsymbol{\phi}$ to encourage them to not diverge from each other. This invalidates the theoretical guarantees about the variance of the positive and negative partitions of $\mathbf{w}$ as defined in equations \eqref{eq:eff_a} - \eqref{eq:eff_e} however it works well in practice and alleviates the need to introduce superfluous saturating nonlinearities. All results discussed in subsequent sections do not include any activations or saturating nonlinearities added to the original full-precision architectures in the forward or backward propagation and instead $L_2$ regularization is applied to $\boldsymbol{\phi}$ to encourage the full-precision parameters to not diverge far from zero.

\subsection{Progressive binarization}
\label{S:Progressive binarization}
Experimental results demonstrate that it might be beneficial to gradually increase the separation between the quantized values during training by interpolating the binarized weights $\mathbf{w}$ and the continuous parameters $\boldsymbol{\phi}$ as such:

\begin{align}
\mathbf{w}_{\alpha} & \coloneqq \alpha \mathbf{w} + \left( 1 - \alpha \right) \boldsymbol{\phi} = \alpha g \left( \boldsymbol{\phi}, \zeta \right) + \left( 1 - \alpha \right) \boldsymbol{\phi} \label{eq:alpha_interpolation} \\ 
\alpha & = \left\{
    \begin{array}{l}
      min(\frac{t}{T_{\alpha} * T}, 1) \qquad T_{\alpha} > 0 \\
      1 \qquad \qquad \qquad \: \, T_{\alpha} = 0 \\
    \end{array}
  \right. \label{eq:alpha_calc} 
\end{align}

Where $t$ is the training step number; $T$ is the total number of training steps and $T_{\alpha} \in [0,1]$ is a hyperparameter denoting the fraction of total training steps required for $\alpha$ to reach and remain at 1.

\begin{figure}[H]
\centering
\begin{subfigure}{.5\textwidth}
    \centering
    \includegraphics[width=1.0\textwidth]{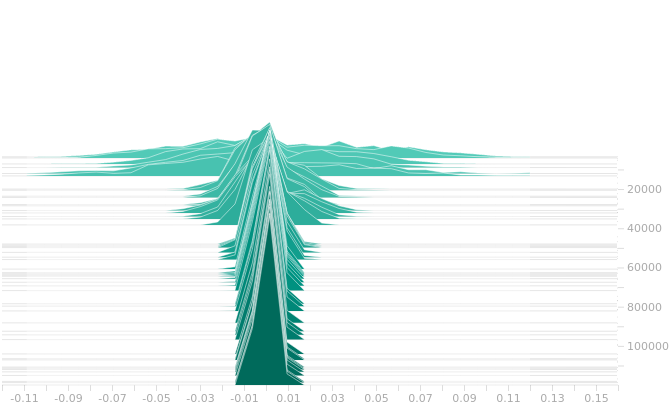}
    \caption{}
\end{subfigure}%
\begin{subfigure}{.5\textwidth}
    \centering
    \includegraphics[width=1.0\textwidth]{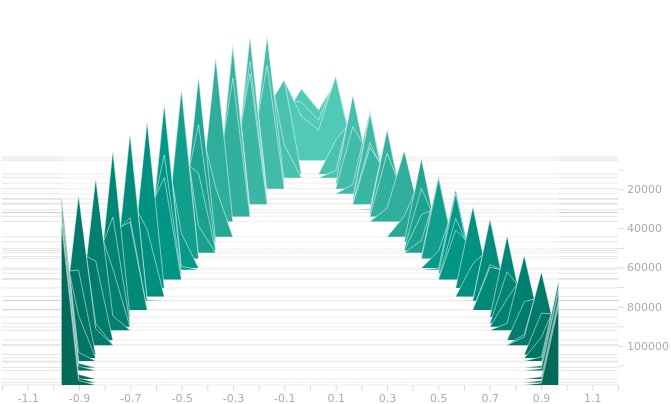}
    \caption{}
\end{subfigure}
\begin{subfigure}{.5\textwidth}
    \centering
    \includegraphics[width=1.0\textwidth]{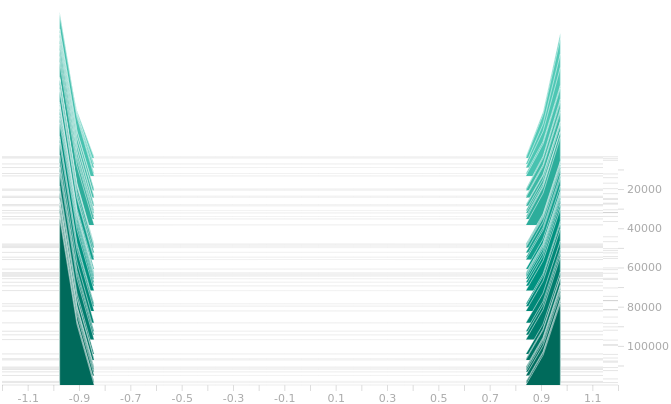}
    \caption{}
\end{subfigure}%
\begin{subfigure}{.5\textwidth}
    \centering
    \includegraphics[width=1.0\textwidth]{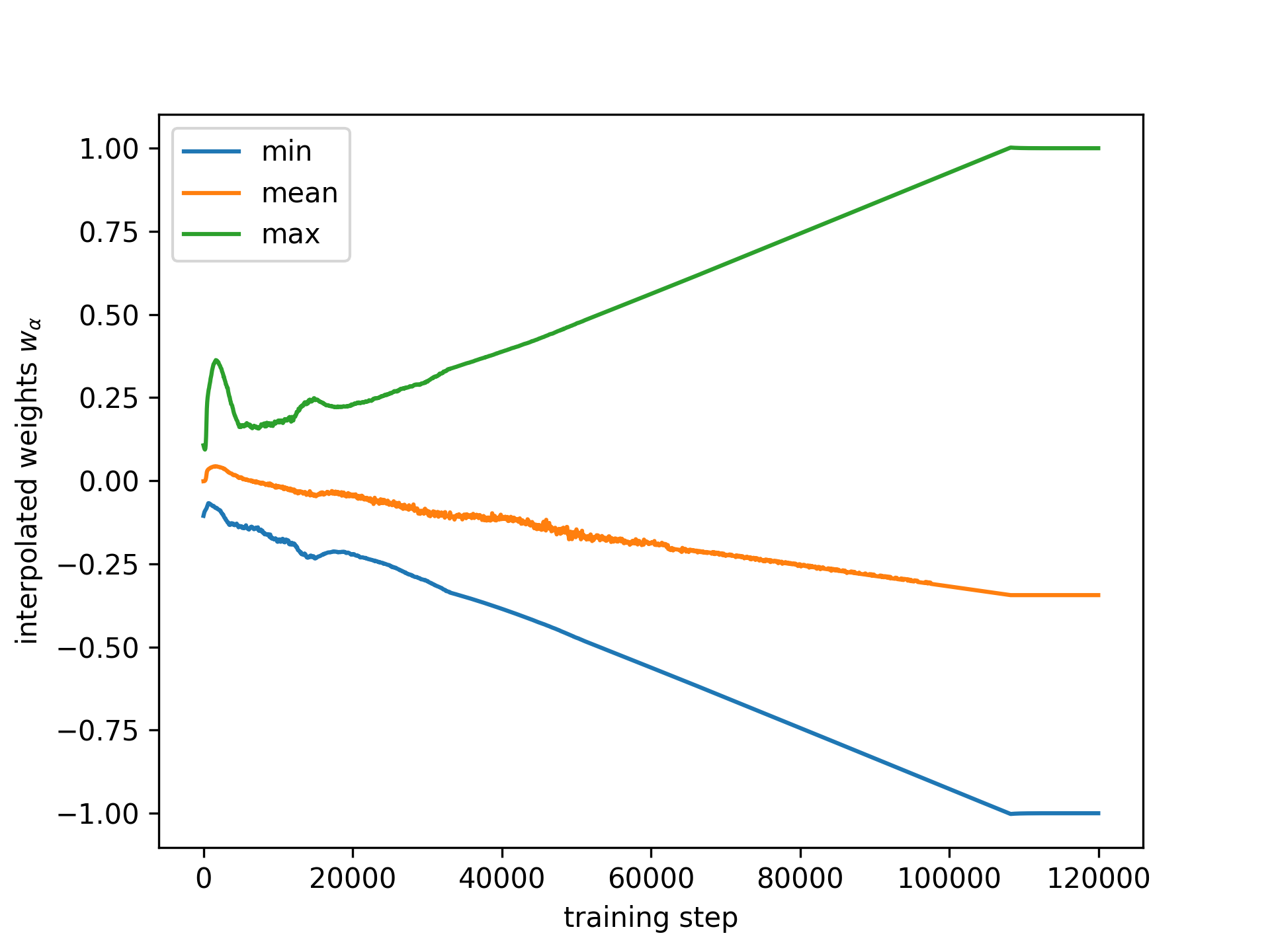}
    \caption{}
\end{subfigure}
\begin{subfigure}{.5\textwidth}
    \centering
    \includegraphics[width=1.0\textwidth]{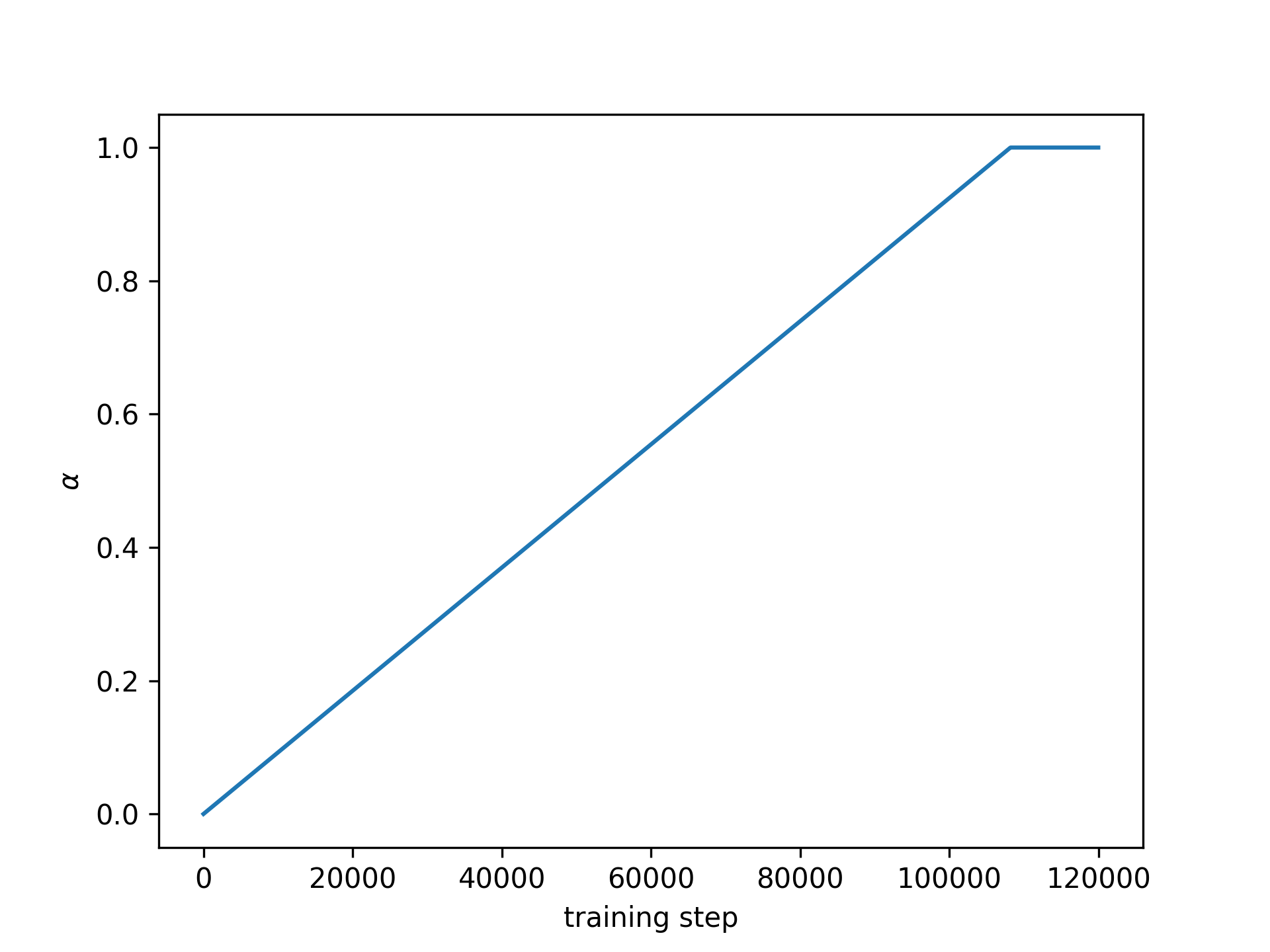}
    \caption{}
\end{subfigure}%
\begin{subfigure}{.5\textwidth}
    \centering
    \includegraphics[width=1.0\textwidth]{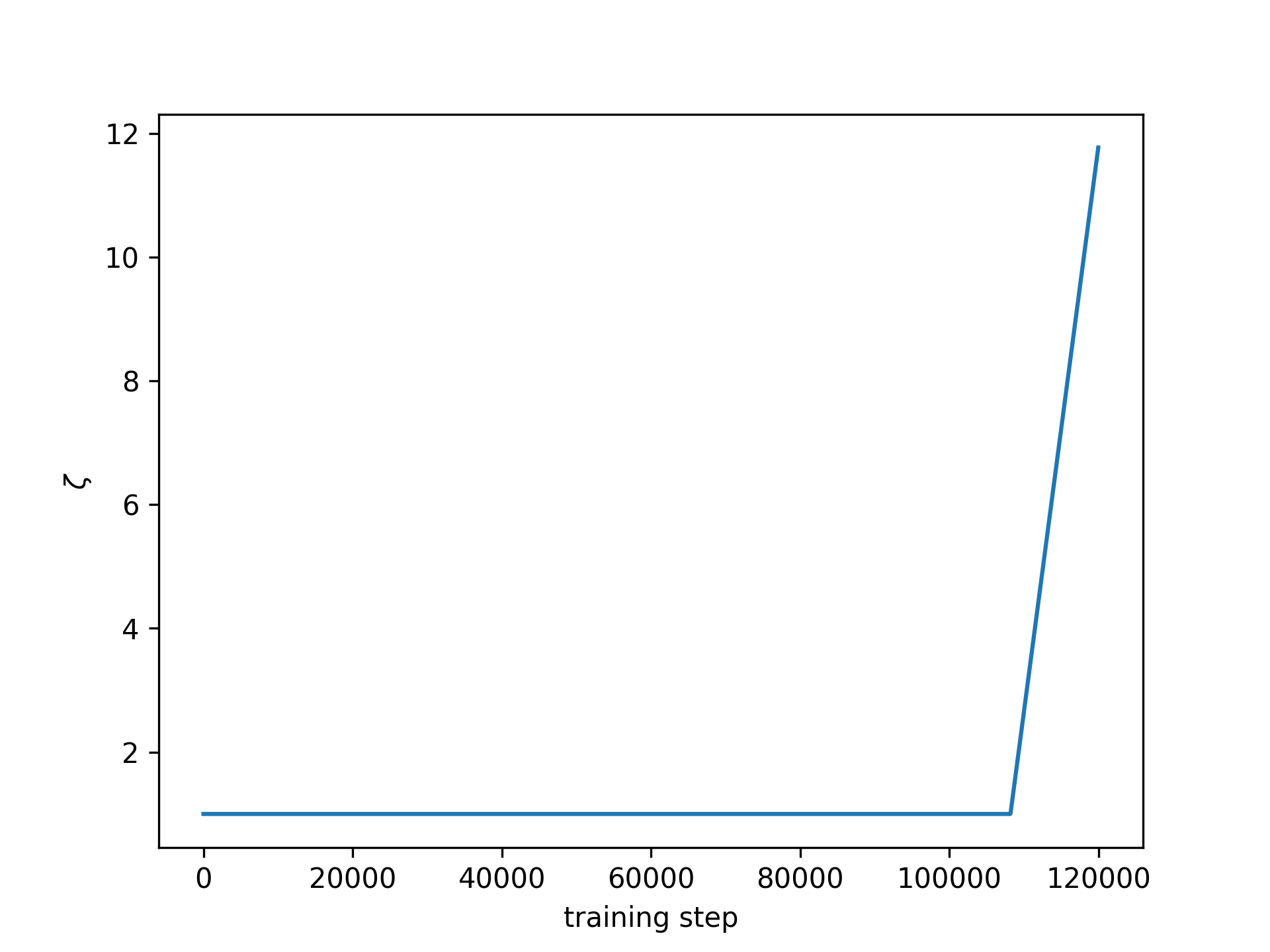}
    \caption{}
\end{subfigure}

\caption[short]{Typical state evolution of a single layer during training when $T_{\alpha}=0.9$. (a) $L_2$ regularized full-precision parameters $\boldsymbol{\phi}$; (b) the interpolated progressively binarized weights $\mathbf{w}_{\alpha}$ equation \eqref{eq:alpha_interpolation}; (c) the weights used in inference simply calculated as $sign(\boldsymbol{\phi})$; (d) the minimum, mean and maximum values of $\mathbf{w}_{\alpha}$ (e) value of $\alpha$ for training step; (f) value of $\zeta$ for training step.}
\label{fig:wrn_plots}
\end{figure}

\subsection{Parameter partitioning}
\label{S:Parameter partitioning}
Partitioning the parameters $\boldsymbol{\phi}$ is useful to limit the breadth of the dependencies introduced by the mean subtraction in equations \eqref{eq:eff_c} and \eqref{eq:eff_d}. Whilst the proposed method supports any arbitrary partitioning scheme, in this work the parameters are partitioned by filter for convolutional layers and by neurons for fully connected layers.

\subsection{Inference}
\label{S:Inference}
In inference the parameters $\boldsymbol{\phi}$ are binarized simply by using the $sign(\boldsymbol{\phi})$ function so that they are restricted to exact values in $\{-1,1\}$. Note that any zero valued parameters are assigned the value -1. At  the completion of training only the binarized weights $\mathbf{w}$ are retained, there is no need to keep the full-precision parameters $\boldsymbol{\phi}$ nor any partitioning related information. 

\section{Related work}
\label{S:RelatedWork}
The proposed method is closely related to the core idea proposed in \cite{shulman2020diffprune} where a similar transformation is used to emulate a multivariate Bernoulli random variable. Whereas in \cite{shulman2020diffprune} nuisance parameters are added to the model to calculate the MLE for multiplicative binary gates in the context of network pruning, in this work no additional parameters are introduced and the weights themselves are transformed to approximate a multivariate binary state over the network weights.

Network quantization refers  to  quantizing  the  weights and/or the activations of an ANN. It is one of a few methods for model compression and efficient model inference and has a large body of work in the literature dedicated to it. The focus of the method proposed in this work is on the extreme scenario of weights binarization to $\{-1,1\}$ offering the maximal compression and speed gains. Since there are too many methods related to BNNs to mention in detail, the interested reader is referred to \cite{journals/corr/abs-1710-09282, Qin_2020, electronics8060661} for a thorough review. The rest of this section is dedicated to methods that solve the binarization problem by smoothing or reinterpreting the combinatorial problem in a way that enables use of exact gradients with backpropagation.

The method proposed in \cite{DBLP:journals/corr/abs-1904-05868} approximates the quantization function $sign(\cdot)$ with $tanh(\cdot)$ such that the estimation error is controlled by gradually scaling the inputs to the quantizer during training. A different approach is taken by \cite{Martinez2020Training} suggesting to train identical networks four times with an alternating teacher-student relationship. An auxiliary loss is added to coerce the networks to learn similar activations. Furthermore they also utilize the hyperbolic tangent function to smooth the $sign(\cdot)$ function. Differentiable Soft Quantization (DSQ) is a method proposed in \cite{Gong_2019_ICCV} to approximate the standard binary and uniform quantization process. DSQ employs a series of hyperbolic tangent functions to form a smooth function that progressively approaches a discrete like state emulating low-bit uniform quantization e.g., $sign(\cdot)$ for the 1-bit case. Continuous Binarization introduced in \cite{SakrCWGS18} approximates the binary activation threshold operation using parameterized clipping functions and scaled  binary activation function. This enables training with exact gradients however the method relies on a custom and lengthy training regime for individual layers and additional regularization. Furthermore the clipping functions are rectified and therefore suffer from zero gradient outside the clip boundaries. Self-Binarizing Networks introduced in \cite{lahoud2019selfbinarizing} approach the binarization task by approximating the $sign(\cdot)$ with hyperbolic tangent which is iteratively sharpened during training. Stochastic Quantization (SQ) \cite{dong2017learning} propose to quantize only a subset of the parameters at a time based on a stochastic selection criteria such that only a subset of the gradients are estimated during backpropagation.

\section{Experiments}
\subsection{Inductive image classification}
To demonstrate the effectiveness of the proposed method the top-1 accuracy is compared between a full-precision architecture and its binary weighted counterpart on a number of inductive image classification tasks. The methodology involves training each model twice, once with full-precision floating point weights and again using the proposed method. Both networks are evaluated at the end of each epoch and the best result achieved on the validation set during training is reported. The models are implemented in TensorFlow \cite{tensorflow2015-whitepaper} using custom Dense and Conv2D layers. The optimizer used in all experiments is the weight decay decoupled SGD momentum optimizer \cite{DBLP:conf/iclr/LoshchilovH19} with a linear learning rate warmup period of 5 epochs. An exponential reduction schedule is applied to both the learning rate and weight decay. In all experiments of full-precision networks, except for the WRN-28-10 CIFAR10, the schedule updates by a factor of 0.1 at 1/3 and 2/3 of the overall post-warmup training steps. For the WRN-28-10 CIFAR10 experiment the schedule updates are as recommended in \cite{BMVC2016_87}. For the binary variants the updates occur at 0.1, 0.25, 0.4, 0.55, 0.7, 0.85 of the overall post-warmup training steps with a factor of 0.3. The parameters of the batch normalization layers are excluded from weight decay. In all experiments $\zeta$ is set to 1 for the initial 90\% of training steps and during the last 10\% of training $\zeta$ is incremented every step until a final value of 12. The training parameters for all experiments are summarized in table \ref{tab:train_params}. The residual blocks all use parameter free identity mapping that downsample skip connections by average pooling and concatenate zeros where required to match the number of activation planes. For the CIFAR data sets classification tasks a basic augmentation of horizontal flip, random translation and zoom is used and in the binary weighted variants all layers are binarized except for the first and last layers of the networks. Note there was no attempt to perform an exhaustive search of hyperparameters for the best possible result therefore these results should be taken as indicative only. All image data sets are taken from TensorFlow Data sets \cite{TFDS} with the default train/test split. The source code is publicly available at \url{https://bitbucket.org/YanivShu/binary_weighted_networks_public}.

\begin{table}[t]
\centering
\begin{tabular}{l l c c c } 
\toprule
Data Set & Architecture & 32b Error \% & 1b Error \% & Change \\ [0.5ex] 
\midrule
MNIST & LeNet5 & 0.64 & 0.53 & 0.11 \\
\hline
\multirow{3}{*}{CIFAR10} & VGG-Small & 5.88 & 6.42 & -0.54 \\
& ResNet-18 & 5.52 & 5.63 & -0.11 \\
& WRN-28-10 & 4.51 & 4.39 & 0.12 \\
\hline
\multirow{2}{*}{CIFAR100} & ResNet-18  & 23.52 & 24.02 & -0.5 \\
& WRN-28-10 & 21.34 & 20.52 & 0.82 \\
\bottomrule
\end{tabular}
\caption[short]{\label{tab:error_rate_results}Summary of experimental results by architecture, data set and weight precision. The error rates are the minimum errors obtained during training for the validation set.}
\end{table}

\begin{table}[t]
\begin{center}
\begin{tabular}{l c c c c c c} 
\toprule
Architecture & Bits & Batch & Epochs & L.R. & W.D. & $T_{\alpha}$ \\ [0.5ex] 
\midrule
\multirow{2}{*}{LeNet5} & 32 & \multirow{2}{*}{100} & \multirow{2}{*}{200} & \multirow{2}{*}{0.01} & 1e-4 & - \\
 & 1 & & & & 1e-3 & 0.9 \\
\hline
\multirow{2}{*}{VGG-Small} & 32 & \multirow{2}{*}{128} & \multirow{2}{*}{300} & 0.1 & 5e-4 & - \\
& 1 & & & 0.05 & 1e-3 & 0 \\
\hline
\multirow{2}{*}{ResNet-18} & 32 & \multirow{2}{*}{128} & 300 & 0.1 & 5e-4 & - \\
& 1 & & 400 & 0.05 & 1e-3 & 0.9 \\
\hline
\multirow{2}{*}{WRN-28-10} & 32 & \multirow{2}{*}{128} & 200/300 & 0.1 & 5e-4 & - \\
& 1 & & 400 & 0.05 & 1e-3 & 0.9 \\
\hline
\end{tabular}
\end{center}
\caption[short]{Summary of hyperparameters for all experiments. Bits is the bit depth of the network weights. Batch is the batch size used in training. Epochs is the total number of training epochs. L.R. is the initial post warmup learning rate. W.D. is the initial weight decay scaler. $T_{\alpha}$ denotes the fraction of total training steps required for $\alpha$ to reach and remain at 1, see equation \eqref{eq:alpha_calc}. The 200/300 in the WRN-28-10 row indicates the number of training epochs for the CIFAR10/CIFAR100 data sets respectively.}
\label{tab:train_params}
\end{table}

\subsection{LeNet5 MNIST classification}
The first experiment is the toy classification task of MNIST using the basic CNN LeNet5 \cite{Lecun98gradient-basedlearning}. In the binary weighted variant all layers except for the last dense prediction layer are binarized.

\subsection{CIFAR10 classification}
The second experiment is the classification task of the CIFAR10 data set with three different architectures: Vgg-Small like network similar to the one used in \cite{10.1007/978-3-030-01237-3_23}, ResNet-18 \cite{7780459} and WRN-28-10 \cite{BMVC2016_87}. For WRN-28-10 the baseline architecture is the no dropout variant with identity mapping. A minor modification was done to the architecture by increasing the number of filters in the first convolution layer from 16 to 64.

\begin{table}[!ht]
    \scriptsize
    \begin{minipage}{.5\linewidth}

      \centering
        \begin{tabular}{l c c} 
\toprule
Method & Architecture & Error \% \\
\midrule
LAB \cite{DBLP:conf/iclr/HouYK17} \cite{Qin_2020} & VGG-Small & 10.5 \\
BWN \cite {DBLP:conf/eccv/RastegariORF16} \cite{Qin_2020} & VGG-Small & 9.9 \\
Self-Binarizing Networks \cite{lahoud2019selfbinarizing} & VGG-Small & 9.4 \\
BWNH \cite{conf/aaai/HuWC18} & VGG9 & 9.2 \\
MPT-1/32 (95) \cite{diffenderfer2021multiprize} & VGG-Small & 8.5 \\
BinaryConnect \cite{NIPS2015_3e15cc11} \cite{Qin_2020} & VGG-Small & 8.3 \\
\textbf{Proposed method} & VGG-Small & \textbf{6.4} \\
\bottomrule
\end{tabular}
    \end{minipage}%
    \begin{minipage}{.5\linewidth}
      \centering
        \begin{tabular}{l c c} 
\toprule
Method & Architecture & Error \% \\
\midrule
IR-Net \cite{DBLP:conf/cvpr/QinGLSWYS20} \cite{Qin_2020} & ResNet-20 & 9.8 \\
ProxQuant \cite{bai2018proxquant} \cite{Qin_2020} & ResNet-20 & 9.3 \\
ProxQuant \cite{bai2018proxquant} & ResNet-44 & 7.8 \\
SQ-BWN \cite{dong2017learning} & ResNet-56 & 7.2 \\
\textbf{Proposed method} & ResNet-18 & 5.6 \\
MPT (80) +BN \cite{diffenderfer2021multiprize} & ResNet-18 & 5.2 \\
\textbf{Proposed method} & WRN-28-10 & \textbf{4.4} \\
\bottomrule
\end{tabular}
    \end{minipage}
    \caption{\label{tab:error_rate_results}Comparison of reported error rates on the CIFAR10 validation set for binary weighted networks. The left table summarizes the results for VGG based architectures and the right table the results for ResNet based architectures. The citations indicate the paper where the method is proposed and the source of the results if different to the paper.} 
\end{table}

\subsection{CIFAR100 classification}
The third experiment is the classification task of the CIFAR100 data set using the ResNet-18 \cite{7780459} and WRN-28-10 \cite{BMVC2016_87} architectures identical to these used in the CIFAR10 experiments. Note that attempting training the full-precision WRN-28-10 network with the same hyperparamter settings and learning rate schedule as specified in \cite{BMVC2016_87} resulted in a slightly reduced accuracy.

\begin{table}[h]
\centering
\begin{tabular}{l c c} 
\toprule
Method & Architecture & Error \% \\ [0.5ex] 
\midrule
Self-Binarizing Networks \cite{lahoud2019selfbinarizing} & VGG-Small & 36.5 \\
BWN \cite{DBLP:conf/eccv/RastegariORF16} \cite{dong2017learning} & ResNet-56 & 35.0 \\
BWNH \cite{conf/aaai/HuWC18} & VGG9 & 34.4 \\
SQ-BWN \cite{dong2017learning} & ResNet-56 & 31.6 \\
\textbf{Proposed method} & ResNet-18 & 24.0 \\
\textbf{Proposed method} & WRN-28-10 & \textbf{20.5} \\
\bottomrule
\end{tabular}
\caption{\label{tab:error_rate_results}Comparison of reported error rates on the CIFAR100 validation set for binary weighted networks. The citations indicate the paper where the method is proposed and the source of the results if different to the paper.} 
\end{table}

\subsection{Effect of $T_{\alpha}$}
This section aims to quantify the effect of progressive binarization with different rates $T_{\alpha}$. For this purpose models are trained a number of times with all settings unchanged except for modifying $T_{\alpha}$. The results summarized in table \ref{tab:error_rate_results_prog_bin} indicate that the models can train well with or without progressive binarization. Despite no strong evidence to support the usefulness of applying progressive binarization it seems that for the deeper residual networks slow progressive binarization did slightly improve accuracy on the validation set.

\begin{table}[t]
\centering
\begin{tabular}{l c c c c c } 
\toprule
Experiment & 0 & 0.3 & 0.5 & 0.7 & 0.9 \\ [0.5ex] 
\midrule
VGG-Small CIFAR10 & \textbf{6.42} & 7.04 & 6.94 & 7.2 & 6.88 \\
ResNet-18 CIFAR100 & 24.61 & 24.78 & 24.81 & 24.56 & \textbf{24.02} \\
WRN-28-10 CIFAR100 & 20.59 & 20.67 & 20.96 & 20.92 & \textbf{20.52} \\
\bottomrule
\end{tabular}
\caption[short]{\label{tab:error_rate_results_prog_bin}Best accuracy measured on the validation set during training for different values of $T_{\alpha}$.} 
\end{table}

\begin{figure}[h]
\centering
\begin{subfigure}{.5\textwidth}
    \centering
    \includegraphics[width=1.0\textwidth]{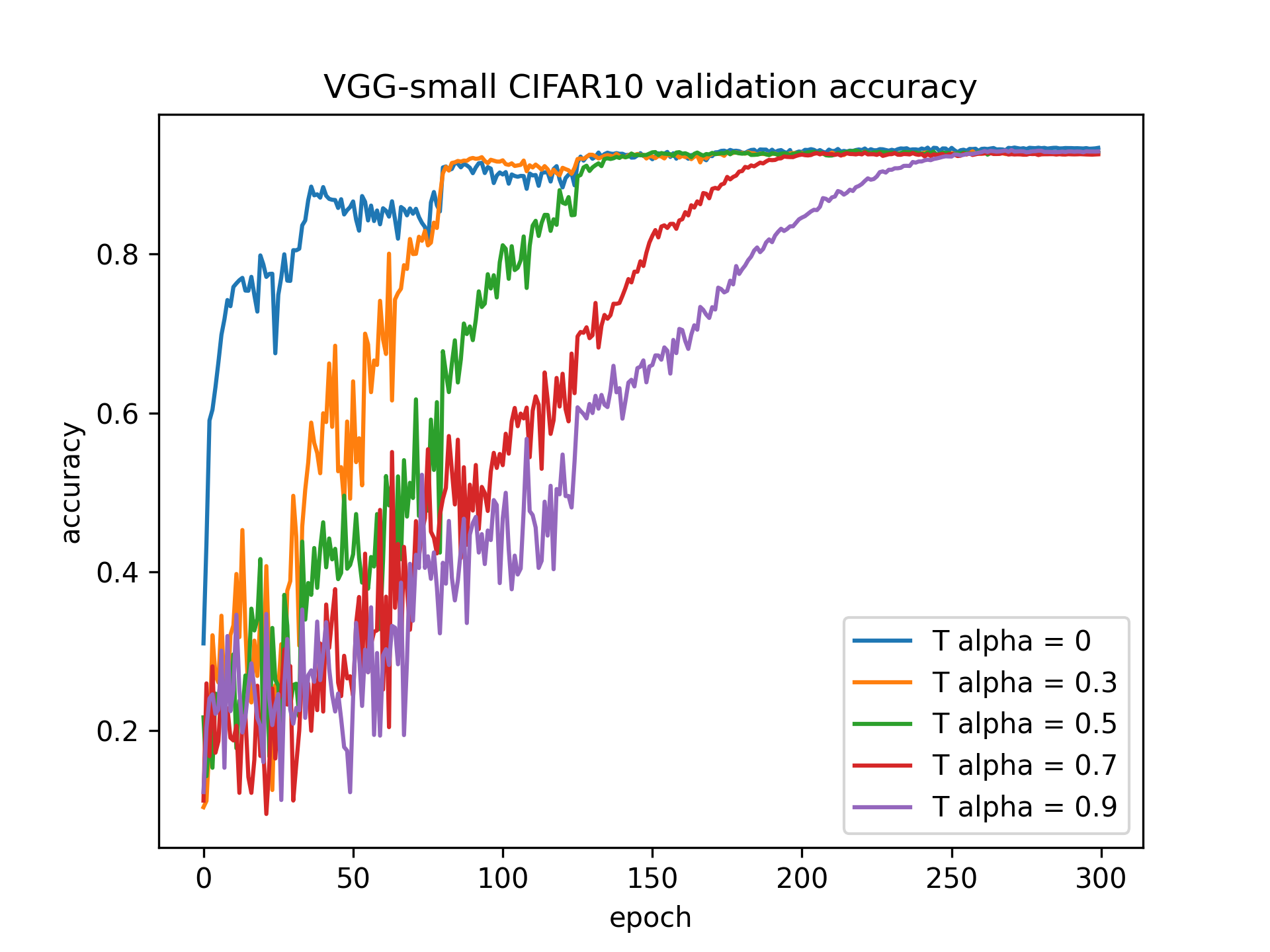}
    \caption{}
\end{subfigure}%
\begin{subfigure}{.5\textwidth}
    \centering
    \includegraphics[width=1.0\textwidth]{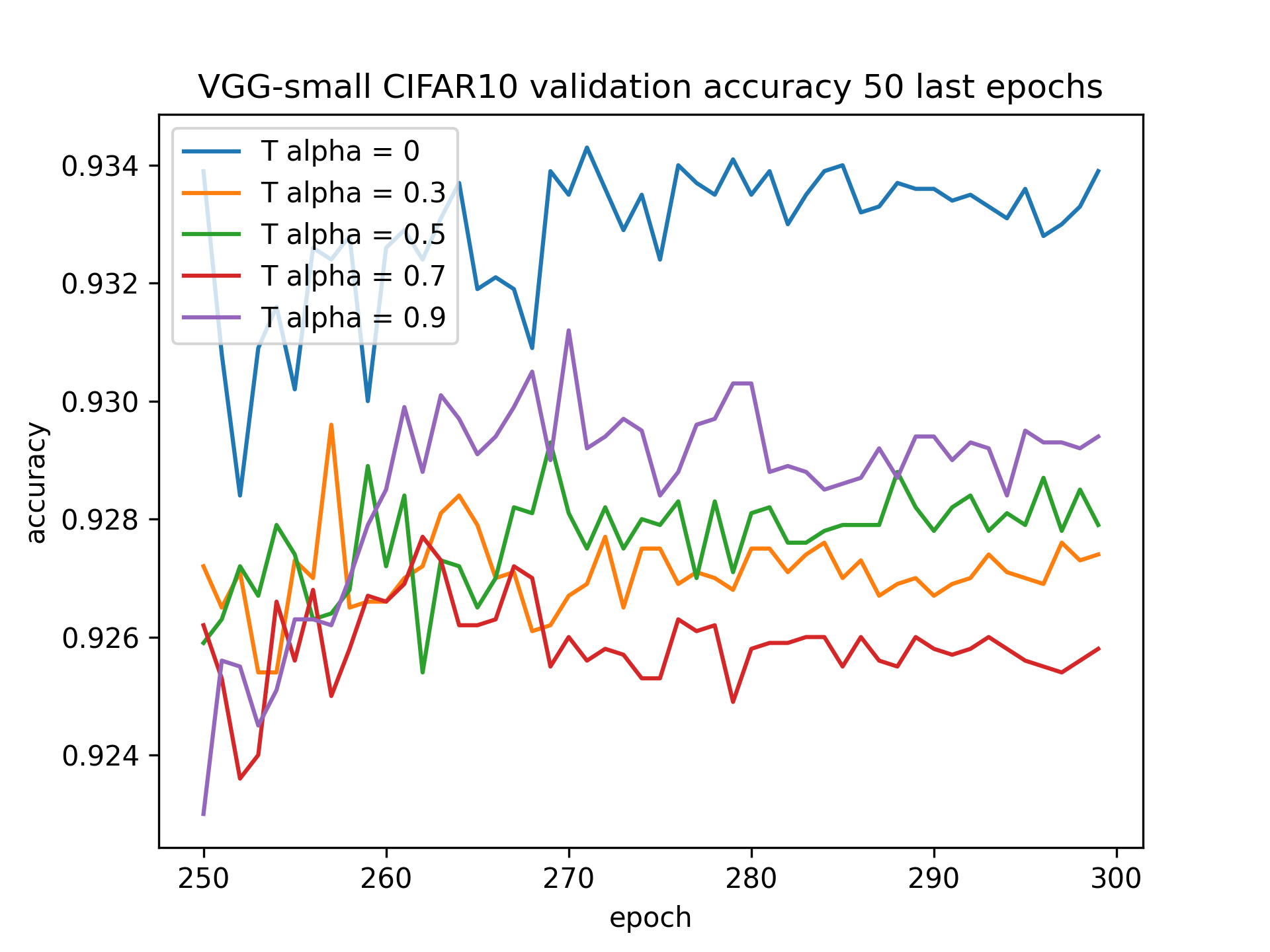}
    \caption{}
\end{subfigure}
\begin{subfigure}{.5\textwidth}
    \centering
    \includegraphics[width=1.0\textwidth]{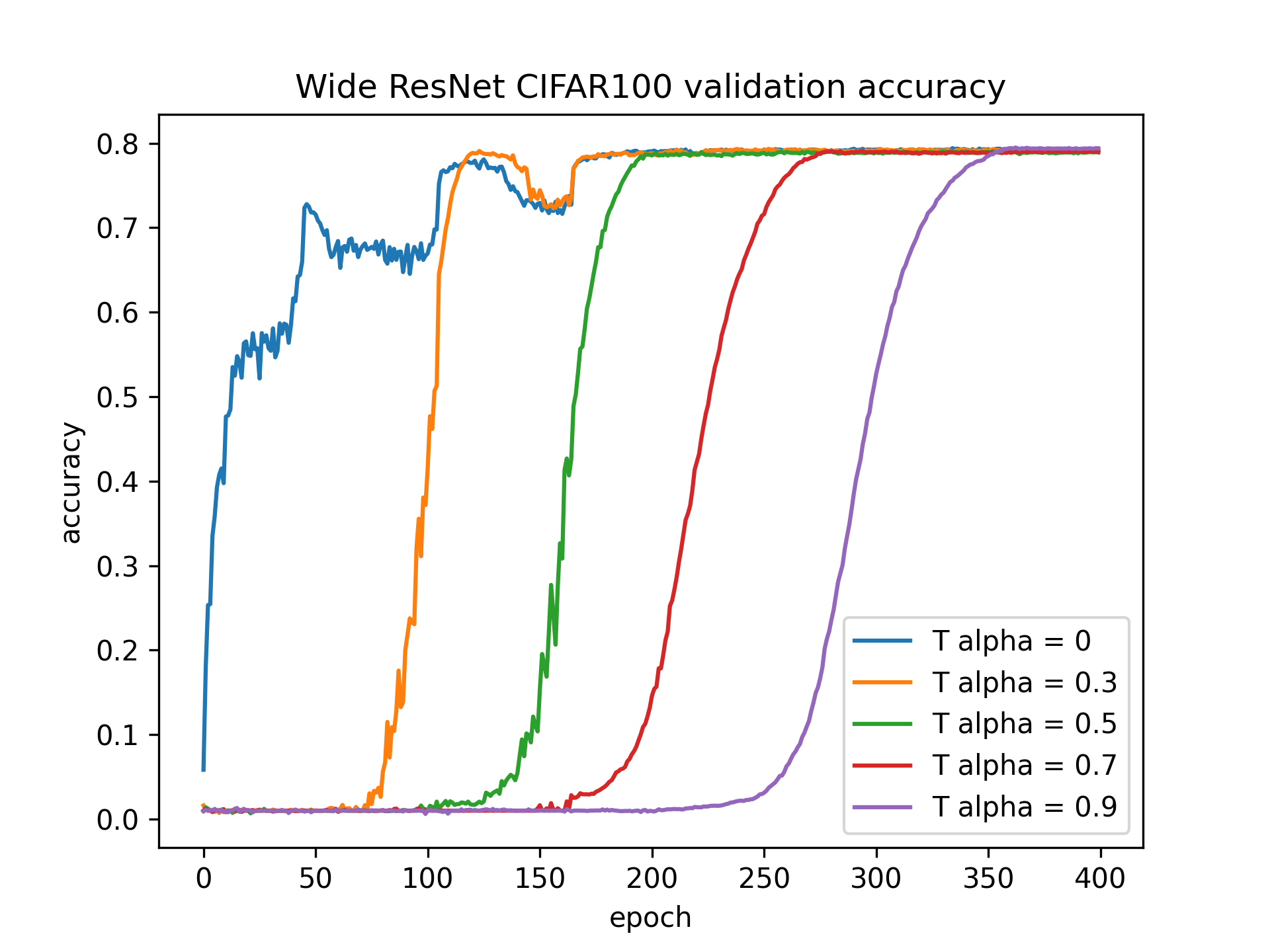}
    \caption{}
\end{subfigure}%
\begin{subfigure}{.5\textwidth}
    \centering
    \includegraphics[width=1.0\textwidth]{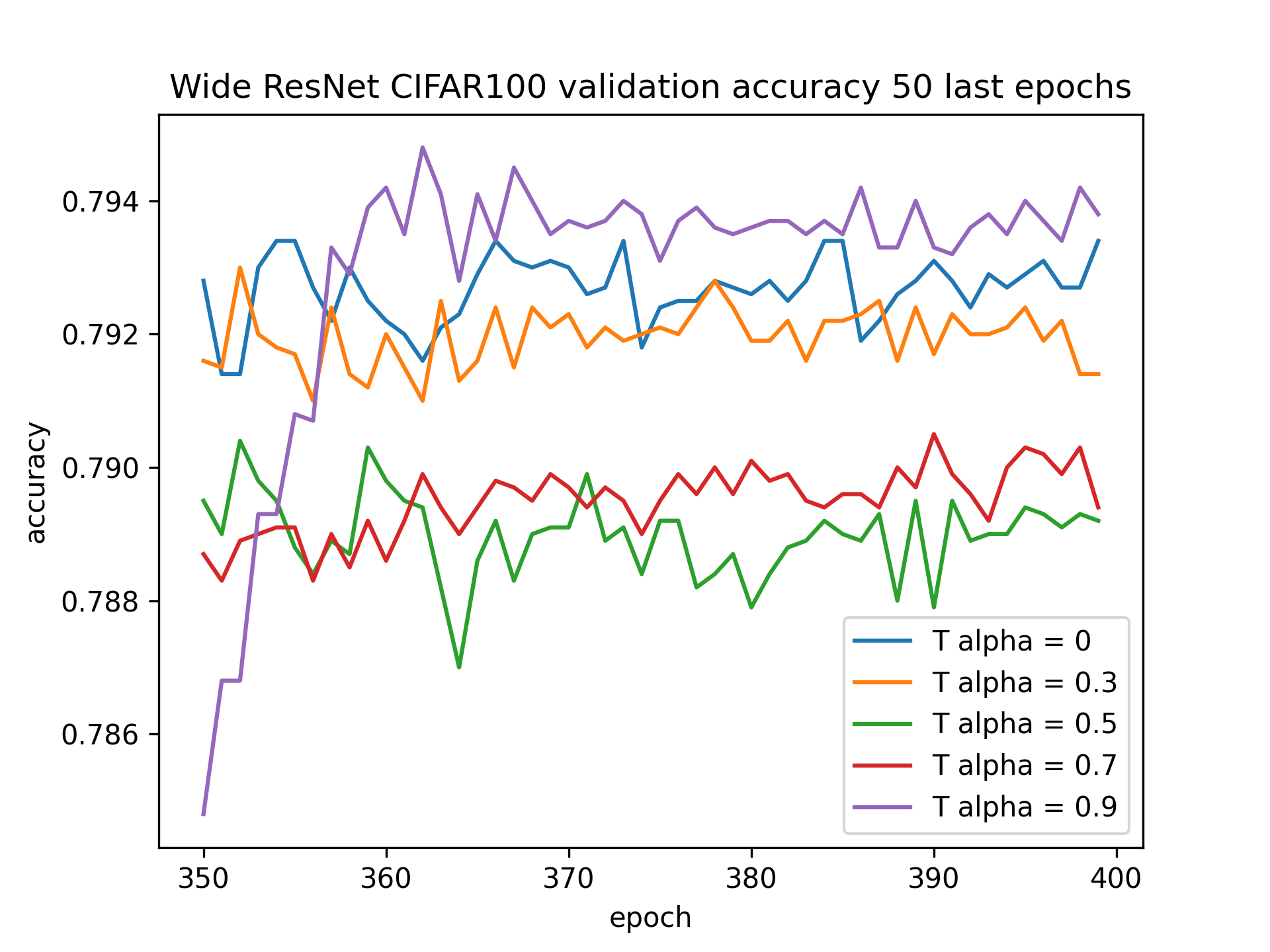}
    \caption{}
\end{subfigure}
\caption[short]{(a, c) Accuracy measured on the validation sets for the VGG-Small CIFAR10 and Wide ResNet CIFAR100 classification tasks for the binary weighted variant with different values of $T_{\alpha}$ over the entire training. (b, d) The same for the last 50 epochs of training.}
\label{fig:wrn_plots}
\end{figure}

\section{Discussion}
In this section an analysis is performed to investigate the reasons leading to the outstanding experimental results. Consider the dot product, the core operation of neural networks, and it's gradient:

\begin{align}
y & = \boldsymbol{\phi}^T \mathbf{x} \\
\dfrac{\partial}{\partial \boldsymbol{\phi}} \sigma (y) & = \dfrac{\partial \sigma}{\partial y} \dfrac{\partial y}{\partial \boldsymbol{\phi}} = \dfrac{\partial \sigma}{\partial y} \mathbf{x} \label{eq:gradient_orig}
\end{align}

Where $\mathbf{x}, \boldsymbol{\phi} \in \mathbb{R}^n$; and $\sigma$ is an arbitrary nonlinearity. In comparison, consider the positive (or negative) group transformation proposed in this work and its gradient:

\begin{align}
z & = \mathbf{w}^T \mathbf{x} \coloneqq \left( \left( \boldsymbol{\phi} - \bar{\phi}  \right) e^{- \zeta} + 1 \right)^T \mathbf{x} \\
\dfrac{\partial}{\partial \boldsymbol{\phi}} \sigma (z) & = \dfrac{\partial \sigma}{\partial z} \dfrac{\partial z}{\partial \mathbf{w}} \dfrac{\partial \mathbf{w}}{\partial \boldsymbol{\phi}} = 
\dfrac{\partial \sigma}{\partial z} \begin{bmatrix}
     1-\frac{1}{n} & -\frac{1}{n} & \ldots &  &  &  -\frac{1}{n} \\
     -\frac{1}{n} &    &   &    &    &  \\
   \vdots  &    &   &    &    &  \vdots  \\
       &    &  \multicolumn{2}{c}{\smash{\raisebox{.5\normalbaselineskip}{\diagdots{8em}{.5em}}}} &    &  \\
      &    &   &    &    & -\frac{1}{n} \\
    -\frac{1}{n} &  & \ldots &  &  -\frac{1}{n} &  1-\frac{1}{n}
  \end{bmatrix} \mathbf{x} \, e^{- \zeta} \\
 & = \dfrac{\partial \sigma}{\partial z} \left( \mathbf{x} - \bar{x} \right) e^{- \zeta} \label{eq:gradient_g}
\end{align}

Where $\bar{x} = \sum{x_i} / n$ is the mean of the elements of the vector $\mathbf{x}$. Equation \eqref{eq:gradient_g} reveals interesting properties of the proposed method. 

The first is that for each of the partitions the gradients are zero centered due to the mean subtraction. This implies that after the gradient update the mean of the parameters will remain unchanged. Assuming the parameters are initialized with zero mean and considering this in conjunction with the $L_2$ regularization this property may have a regularizing effect. If a probabilistic interpretation is assumed similar to \cite{DBLP:conf/cvpr/QinGLSWYS20}, maintaining the parameters having a close to symmetric distribution with zero mean may increase the entropy of the weights distribution and therefore the representation power of the network.

Secondly, assume that $\bar{x} \approx 0$ then the gradients of the full-precision and binary networks are proportional i.e. $e^{- \zeta} \nabla \sigma (y) = \nabla \sigma (z)$ with equality when $\zeta = 0$. For standard gradient descent the proportionality implies the two models can be trained identically simply by scaling the learning rate. The assumption of $\bar{x} \approx 0$ is reasonable for inputs that are normalized by methods such as batch normalization \cite{10.5555/3045118.3045167}, instance normalization \cite{DBLP:journals/corr/UlyanovVL16} or group normalization \cite{citeulike:14571032}. Therefore training of approximate binary weighted networks with gradient descent can be as effective as the training of full-precision networks as long as the two aforementioned conditions are maintained. 

\section{Conclusion}

This paper proposes a novel and effective method for training binary weighted networks by smoothing the combinatorial problem of finding a binary vector of weights to minimize the expected loss for a given objective by means of empirical risk minimization with backpropagation. The method adds little computational complexity and can be readily applied to common architectures using automatic differentiation frameworks. Theoretical analysis and experimental results demonstrate that binary weighted networks can train well with the same standard optimization techniques and similar hyperparameter settings as their full-precision counterparts such as momentum SGD with large learning rates and $L_2$ regularization.

\bibliographystyle{abbrv}

\bibliography{refs}

\end{document}